\documentclass[lettersize,journal]{IEEEtran}
\usepackage{amsmath,amsfonts}
\usepackage{algorithmic}
\usepackage{array}
\usepackage[caption=false,font=normalsize,labelfont=sf,textfont=sf]{subfig}
\usepackage{textcomp}
\usepackage{stfloats}
\usepackage{url}
\usepackage{verbatim}
\usepackage{graphicx}
\usepackage{subfiles}
\usepackage{adjustbox}
\usepackage{multirow}
\usepackage{multicol}
\usepackage{caption}
\usepackage[normalem]{ulem}
\usepackage{makecell}
\usepackage{xcolor}
\usepackage{booktabs}
\usepackage{tabularx}
\usepackage[flushleft]{threeparttable}
\usepackage{hyperref}
\def\BibTeX{{\rm B\kern-.05em{\sc i\kern-.025em b}\kern-.08em
    T\kern-.1667em\lower.7ex\hbox{E}\kern-.125emX}}
\usepackage{balance}
\begin{document}

\title{PCFEx: Point Cloud Feature Extraction for Graph Neural Networks}

\author{\IEEEauthorblockN{Abdullah Al Masud$^{\text{1}}$, Shi Xintong$^{\text{2}}$, Mondher Bouazizi$^{\text{3}}$, Ohtsuki Tomoaki$^{\text{4}}$}\\
\thanks{
\IEEEauthorblockA{$^{\text{1, 2}}$\textit{Graduate School of Science and Technology, Keio University, Kanagawa, 223-8522, Japan}}\\
\IEEEauthorblockA{$^{\text{3, 4}}$\textit{Faculty of Science and Technology, Keio University
Kanagawa, 223-8522, Japan}}\\
\and
Email: $^{\text{1}}$abdullahalmasud.keio@gmail.com, $^{\text{2}}$shixintong@ohtsuki.ics.keio.ac.jp, $^{\text{3}}$bouazizi@ohtsuki.ics.keio.ac.jp, \\$^{\text{4}}$ohtsuki@keio.jp.\\
Manuscript submitted on June, 2025}
}
\maketitle

\markboth{Journal of IEEE IoT,~Vol.~xx, No.~x, April~2025}{PCFEx: Point Cloud Feature Extraction for Graph Neural Networks}
\begin{abstract}
Graph Neural Networks (GNN) have gained significant attention for their effectiveness across various domains. This study focuses on applying GNN to process 3D point cloud data for Human Pose Estimation (HPE) and Human Activity Recognition (HAR). We propose novel point cloud feature extraction techniques to capture meaningful information at the point, edge, and graph levels of the point cloud by considering point cloud as a graph. Moreover, we introduce a GNN architecture designed to efficiently process these features. Our approach is evaluated on four most popular publicly available millimeter-wave radar datasets—three for HPE and one for HAR. The results show substantial improvements, with significantly reduced errors in all three HPE benchmarks, and an overall accuracy of 98.8\% in mmWave-based HAR, outperforming existing state-of-the-art models. This work demonstrates the great potential of feature extraction incorporated with GNN modeling approach to enhance the precision of point cloud processing.
\end{abstract}

\section{Introduction}

Human Pose Estimation (HPE) and Human Activity Recognition (HAR) have achieved remarkable success through deep learning techniques. Two-dimensional (2D) human pose estimation (HPE) from RGB images is a well-studied topic, with notable works such as \cite{hpe-part-affinity, hrnet, openpose, higher-hrnet}. In 3D HPE, notable progress includes end-to-end models \cite{one-stage-1, one-stage-2, one-stage-4, one-stage-3} and two-stage "lifting" approaches \cite{two-stage-3, two-stage-2, two-stage-1}. However, these RGB-based methods are sensitive to lighting, body orientation, shadows, etc., which degrades performance in HPE and human activity recognition (HAR) and raises privacy concerns in monitoring applications. By contrast, mmWave radar provides reliable 3D data and is robust to visual conditions, leading to better accuracy in HPE, as shown in [11]. It also consumes significantly less power and better preserves privacy. Motivated by these factors, we adopted mmWave radar for our study.

Early works in radar-based HPE such as \cite{rfpose3d} employed CNN-based region proposal networks where \cite{mmpose} employed a forked-CNN to stack radar point clouds, \cite{mars}, Fast-Scalable \cite{fast-scalable}, and mRI \cite{mri}, used similar CNN-based techniques to predict human pose keypoints. In contrast, \cite{mi-mesh} used transformer model for HPE, \cite{mmfi} applied a point-transformer model as well, \cite{hupr} used transformer to predict HAR. \cite{radhar} was the first to combine voxelized radar point clouds with Long Short-Term Memory (LSTM) networks for activity recognition. Further works like \cite{mmpoint-gnn}, introduced GNN for HAR, while \cite{improving-har-gnn} used ST-GCN for sequential pose keypoint input to predict HAR.

However, current approaches still face limitations. Notably, the internal structure of the point cloud is often ignored, requiring models to implicitly learn it resulting in susceptibility towards noisy points leading to poor model convergence. Additionally, in methods like \cite{mmpose} and \cite{mars}, the stacked image array representation of radar data assumes adjacent points are spatially close, distorting the relationships between points in the radar point cloud. The damage in spatial coherency affects the efficacy of the underlying CNN-kernels resulting in poor keypoints prediction

To address these challenges, we propose a GNN-based approach using Graph Attention (GAT) \cite{gat} to effectively combine node and edge features. We introduce \(Statbox\) for node and frame feature extraction, shared Multi-Layer Perceptrons (MLP) for node feature processing, and by proposing frame feature processing block. We validate our method on four most popular mmWave point cloud datasets— MMFi, mRI, MARS, and MMActivity to provide strong empirical support for our hypothesis in both HPE and HAR.

Going forward, we discuss the research background in section II, explain modeling approach in section III, describe implementation details in section IV, discuss the results, insights, improvements in section V and VI and finally wrap up our discussion in section VII and VIII.

\section{Related Work}
Recent works on HPE and HAR with mmWave radar vary widely in how they handle spatial structure, temporal dynamics, and data representations such as point clouds or radar heatmaps. These differences are directly reflected in the underlying model architecture. Consequently, to better present the prior works, we categorize existing methods below by their core neural architectures

\subsubsection{Convolutional neural networks (CNN)}
Recent advancements in HPE and HAR using mmWave radar largely rely on Convolutional Neural Networks (CNNs) to extract spatial and temporal features. \cite{rfpose3d} introduced voxelized 4D tensors (3D space + time) and applied 4D convolutions to predict 14 body keypoints. \cite{mmpose} projected 3D radar point clouds onto 2D planes (XY, YZ) and used a forked-CNN, while \cite{mars} stacked point clouds into image-like tensors for CNN-based pose estimation. \cite{fast-scalable} and \cite{mri} enhanced frame density through point cloud fusion. Temporal dynamics were modeled in \cite{joint-global-local} and \cite{shi} via CNN-LSTM combinations over range-Doppler maps. \cite{throughwall-xformer} used CNNs and transformers \cite{xformer} on raw radar signals for 3D pose estimation. \cite{fast-rfpose} localized human regions with CNN before keypoint prediction; \cite{xformer-hpe} combined 3D convolutions and self-attention for 2D pose estimation. In HAR, \cite{radhar}, \cite{noninvasive-har} employed CNN-LSTM models, and \cite{fast-har} integrated 3D CNNs with self-attention. However, methods like \cite{mmpose, mars, mri, fast-scalable} rely on image-like tensor representations, distorting spatial relationships among radar points and limiting CNN effectiveness due to their dependence on spatial locality. It also makes them vulnerable to signal noises. Others—\cite{joint-global-local, throughwall-xformer, fast-rfpose, xformer-hpe}—use radar heatmaps that often exclude dimensions like elevation, Doppler velocity, and intensity. In contrast, our method processes raw 3D point clouds directly as graphs, preserving spatial structure. By leveraging GAT, we employ a more robust architecture against signal noises, we extract rich node, edge, and frame-level features, enabling more tailored towards accurate and structurally aware HPE and HAR.

\subsubsection{Multi layer perceptrons (MLP)}
Some researchers have used shared MLP networks for processing 3D point cloud data. \cite{mm-mesh} employed MLP to process raw point clouds and incorporated LSTM \cite{lstm} for handling sequential frames. Similarly, \cite{joint-dynamic} utilized MLP akin to PointNet \cite{pointnet} and added LSTM for sequence modeling. A comparable approach was adopted in \cite{constrained-mmwave-hpe}, where spatial and temporal constraints were added to PointNet and LSTM for sequential HPE. \cite{mmskeleton} followed a similar approach, combining PointNet with spatial attention, LSTM, and temporal attention for predicting sequential HPE.
However, these MLP-based models treat each point independently, failing to capture the local spatial relationships that are crucial in point cloud data with strong geometric correlation. While our method also leverages shared-MLPs for initial node and edge feature extraction, it integrates them with GAT to effectively model inter-point relationships, enabling a more expressive and tailored feature representation.

\subsubsection{Transformers}
The transformer architecture \cite{xformer} has gained popularity due to its ability to extract features through self-attention and cross-attention. Researchers have applied transformers for radar-based pose estimation. \cite{mi-mesh} used a transformer encoder to process radar features extracted by PointNet++ \cite{pointnet++} and image features processed by 2D CNNs to predict 3D human mesh. \cite{hupr} employed a transformer encoder to process VRDAE (velocity, range, Doppler, azimuth, elevation) 3D tensors to predict 2D poses, and refined predictions with a GNN. \cite{mmfi} applied the Point-Transformer \cite{point-xformer} for 3D pose estimation from radar Because of radar heatmaps, these approaches still lack complete information such as Doppler velocity, signal intensity, SNR etc. In \cite{hupr}, VRDAE maps were used to capture more information, but this increases computational complexity. To reduce complexity, we introduce grid-based node sampling for dense point clouds, nearest neighbor based edge sampling, along with edge and frame feature extraction to capture the internal state of the point cloud.

\subsubsection{Graph neural networks (GNN)}
Graph Neural Networks (GNN) have also been used for point cloud data processing in HPE and HAR. \cite{improving-har-gnn} used ST-GCN \cite{st-gcn} for sequential radar point clouds processing in HAR. \cite{mmpoint-gnn} applied GNN for HAR, representing the point cloud as a graph, similar to our approach. However, these approaches lack in modeling the relationship inside point-cloud leading to susceptibility towards noisy points resulting in suboptimal accuracy. Our model counters it by introducing Statbox for node, edge and frame features, we employ GAT for more intuitive accumulation of nodes and edge features which also makes it robust against signal noises.

We propose PCFEx (Point Cloud Feature Extraction) as a complete architecture to process all available data of a point cloud in an efficient way using GNN. This work is an extension of our previous work \cite{mmgat-original}. The notable contributions in this study over the previous one are as follows:
\begin{enumerate}
    \item Existing radar point feature dimensions typically range from 3 (\textit{x, y, z} for object-based clouds in \cite{shapenet}) to 5 \textit{(x, y, z, Doppler velocity, signal intensity)}. We are the first to introduce \textit{node feature processing} and \(Statbox\) shown in Fig. \ref{node-feature-extraction}. Together they increase the point feature dimension from 5D to 19D with scope to extend further.
    \item We propose frame-level feature extraction using \(Statbox\). Node, edge, and frame features together, create enriched feature representation for the point cloud.
    \item We propose a frame-feature processing branch to accommodate frame-level features in the GNN model.
    \item We take the motivation from \cite{pointnet} to process node and edge features with shared-MLP, we use edge feature extraction and graph attention the same way like \cite{mmgat-original}. We incorporate these two approaches into our feature extraction processes and modeling techniques.
    \item We introduce node downsampling to reduce computational complexity in case the point cloud is dense.
    \item We use four publicly available mmWave radar benchmark datasets for more reliable empirical validation.
\end{enumerate}

\section{Proposed Method}
\begin{figure}[t]
\centering\includegraphics[width=.49\textwidth]{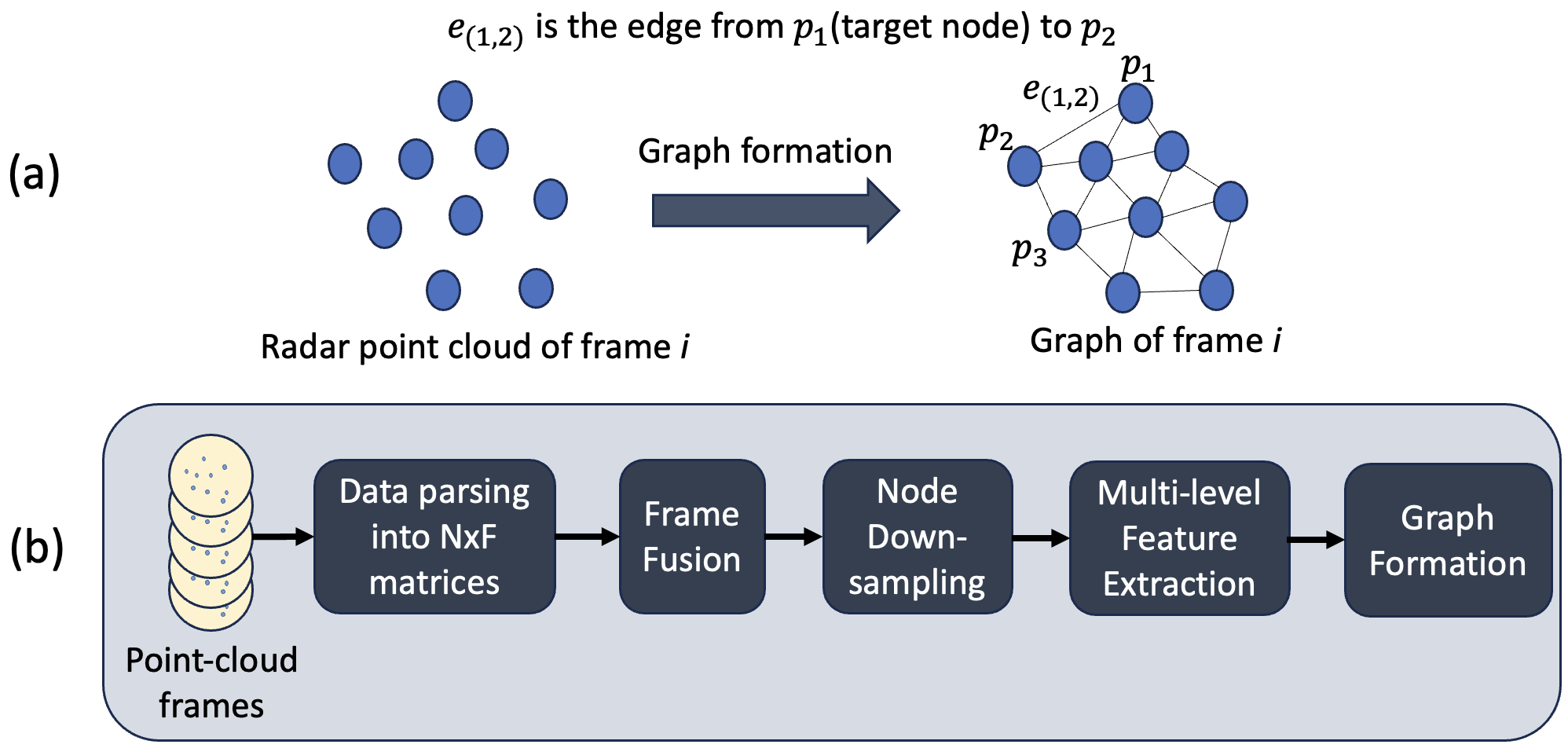}
\caption{(a) point cloud graph formation of frame \(i\); (b) data-processing pipeline.}
\label{graph}
\end{figure}

\subsection{Definition}
Let a radar frame \(i\) contain \(n_i\) data points. We construct a directed graph \(G_i(V, E)\) for frame \(i\), where \(V=\{p_1, p_2,\cdots, p_{n_i}\}\) is the set of nodes and \(E=\{e_{(1,2)}, e_{(2,1)}, \cdots, e_{(n_i,n_i-1)}\}\) is the set of directed edges. Connection from \(p_1\) to \(p_2\) is denoted by \(e_{(1,2)}\) when \(p_1\) is the target node and \(p_2\) is a neighboring node of \(p_1\). The terms ``node" and ``point" are interchangeably used to indicate a single radar data point. A target point \(p_j\) is connected to \(K\) nearest points \(N(p_j)\) based on Euclidean distance such as \(N(p_j)\) = \(\{p_3, p_7, p_8,\cdots, p_{n_i-5}\}\). Therefore, \(e_{(p_j,p_3)}, e_{(p_j,p_7)},\cdots, e_{(p_j,p_{n_i-5})}\) will be included in \(E\).

\subsection{Data Processing}
\subsubsection{Frame Fusion}
We adopt this approach from \cite{fast-scalable} to merge \(F\) consecutive frames point clouds to increase the point cloud density. According to this, we take the point clouds from \((i-F+1)\) to \(i\), merge all points and place them in frame \(i\).

\subsubsection{Node Downsampling}
We use constant cell size along \(x, y, z\) axes, divide the point cloud into necessary number of grid cells to cover the whole point cloud, then randomly select maximum \(Q\) points from each of the cells containing data points. Cell size may vary along \(x, y, z\) axes.

\subsubsection{Node Feature Extraction}

\begin{figure*}[tb]
\centering
\captionsetup{justification=centering}
\includegraphics[width=1.02\textwidth]{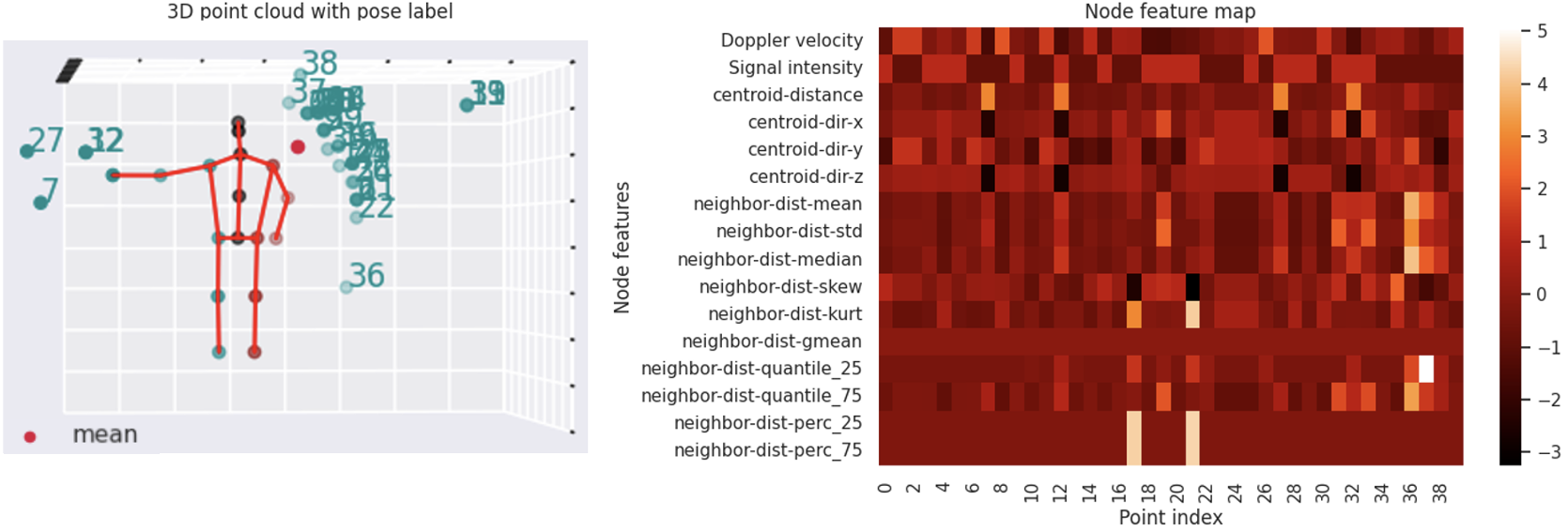}
\caption{Node feature map (right) of a 3D point cloud (left). Points 7, 12, 27, 32 show large centroid distances; point 36 may be an outlier due to high neighbor distance, standard deviation.}
\label{feature-map}
\end{figure*}

\begin{figure}[t]
\centering\includegraphics[width=.48\textwidth]{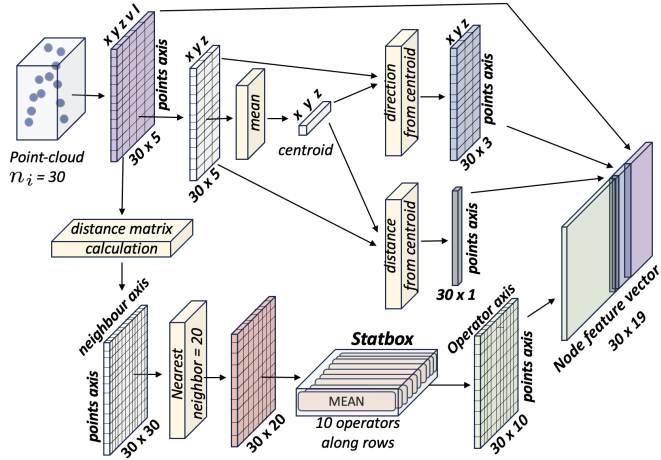}
\caption{Node feature extraction.}
\label{node-feature-extraction}
\end{figure}
In all four datasets MARS\cite{mars}, mRI\cite{mri}, MMFi\cite{mmfi}, MMActivity\cite{radhar}, each point \(p_j\) contains 5 attributes which are \(x, y, z\) coordinates in 3D, Doppler velocity (\(v\)), intensity (\(I\)). We extract additional node features from the \(x,y,z,v,I\) just as shown in Fig. \ref{node-feature-extraction}. Even though the features are  generated from original data, they provide additional information such as the relative position of a point against the neighbors, against the whole point-cloud, distance \& direction from cloud center etc. relatable features regarding point-clouds aiding the model to learn the target task. Their contributions towards the target tasks are well discussed in subsection VI(A). We use \(Statbox\) to extract statistical features which consists of 10 statistical operators (mean, standard deviation, median, skewness, kurtosis, geometric-mean, quantile at 25\% and 75\%, percentile at 25\% and 75\%). We calculate distance and directions from each point to the centroid of the point cloud. Combining all these features with the original \(x,y,z,v,I\), we get node feature \(f_{p_j}\) for point \(p_j\) with 19 dimensions. In Fig. \ref{node-feature-extraction} we show the feature extraction for \(K=8\) for cleaner matrix presentation. However, in all of our experiments, we use \(K=20\) based on the empirical insights from \cite{mmgat-original}.

\begin{figure}[t]
\centering\includegraphics[width=.435\textwidth]{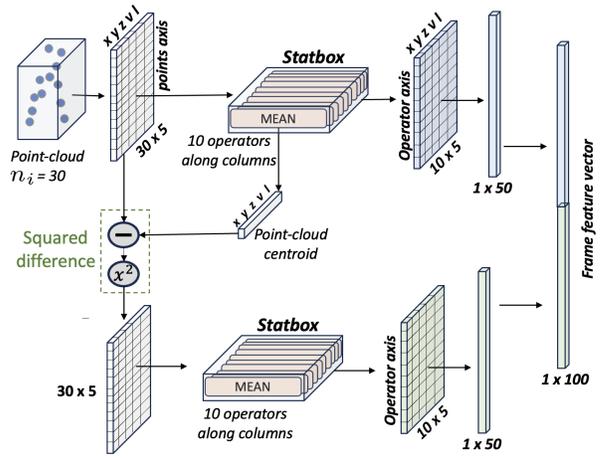}
\caption{Frame feature extraction.}
\label{frame-feature-extraction}
\end{figure}

\subsubsection{Edge Feature Extraction}
For edge feature extraction, we select \(K\) nodes around a target node using nearest neighbor approach. We adopt the edge feature extraction from \cite{mmgat-original} to get 6 features for each selected edge which are: the Euclidean distance  \(||p_1 - p_2||\); the angles from \(p_1\) to \(p_2\), respectively along \(x, y, z\) axes calculated by \((p_2-p_1) / ||p_2-p_1||\), If the denominator is \(0\) the angle is considered as \(0\); the relative velocity  \(v_{p_2}-v_{p_1}\); the relative intensity  \(I_{p_2}-I_{p_1}\). Here \(||.||\) indicates \(L_2\) norm. This 6 dimensional edge feature vector from \(p_1\) towards \(p_2\) is denoted as \(g_{e_{(1,2)}}\).

\subsubsection{Frame Feature Extraction}
Node and edge features computation brings specific information regarding every individual points. However, they do not provide overall information regarding the whole point-cloud which motivated us to design a parallel branch which includes frame feature extraction and processing blocks. They often perform as a residual block by providing a shortcut path for primitive features as described more in subsection VI(A). We extract frame features as shown in Fig. \ref{frame-feature-extraction}. Using \(Statbox\), we compute statistics along the columns (number of samples) of the input point cloud matrix, which also gives us the centroid (Mean from \(Statbox\)). We then calculate the squared distance between all points of point cloud and the centroid, apply \(Statbox\) again to extract features along columns. After flattening and concatenating both feature vectors, we obtain the frame feature vector denoted by \(f_i\). For simplicity, we present node feature dimensions as 5 in Fig. \ref{frame-feature-extraction}, but in practice, we calculate frame features after node feature extraction (19 dimensional vector), resulting in a 380-dimensional frame feature vector.

To optimize the computation overhead, we calculate the distance matrix from the squared-distances shown in Fig. \ref{frame-feature-extraction} and used it in node feature extraction and edge feature extraction as well as to define the edge-matrix for the graph. At this point, we have node features, edge features, and frame features. We now define the graph, assign pose or activity label accordingly for HPE and HAR and store the graph for model training and validation.

\subsection{GNN Model Architecture} \label{gnn-model-desc}
\begin{figure*}[tb]
\centering
\captionsetup{justification=centering}
\includegraphics[width=\textwidth]{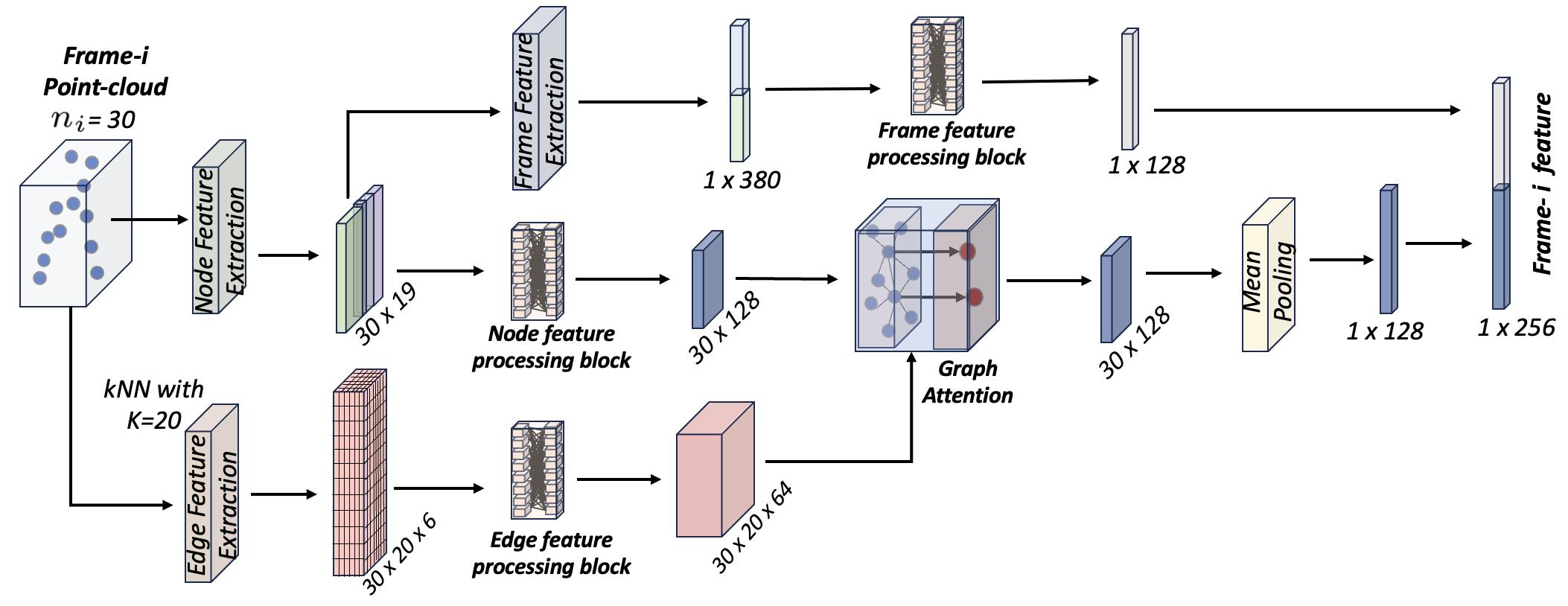}
\caption{Proposed graph neural network. This architecture calculates the frame representation feature vector.}
\label{frame-representation}
\end{figure*}
The whole model consists of two parts. Frame representation feature vector generation shown in Fig. \ref{frame-representation} and prediction head shown in Fig. \ref{prediction-heads}. The frame representation block computes a single feature vector that represents the entire frame, it is then fed into the prediction heads for final prediction.

\subsubsection{Edge Feature Processing Block (\(h_{edge}\))}
The edge feature processing block \(h_{edge}\) consists of shared Fully Connected Network (FCN) layers, each FCN layer is followed by a ReLU \cite{relu} activation, except for the first layer. The processed edge features between a target point \(p_j\) and all its neighbors \(N(p_j)\) are computed like below-
\begin{equation}
    X_{p_{j_{edge}}} = h_{edge}(g_{e_{(p_j,p_k)}}) , p_k \in N(p_j), j \in {1,2,...,n_i}.
    \label{edge-feature}
\end{equation}
It works as a shared-MLP block on all edge features in \(E\).

\subsubsection{Node Feature Processing Block (\(h_{node}\))}
Node feature processing block also consists of FCN layers each followed by ReLU activation. This block also acts as a shared-MLP block on all node features in \(V\). Processed node features \(X_{p_{j_{node}}}\) are calculated as follows:
\begin{equation}
    X_{p_{j_{node}}} = h_{node}(f_{p_j}), j \in {1,2,...,n_i}.
    \label{node-feature}
\end{equation}

\subsubsection{Graph Attention Block}
We combine processed node features \(X_{p_{j_{node}}}\) with the processed edge features \(X_{p_{j_{edge}}}\) in GAT \cite{gat} computation. GAT calculates attention weights for each neighbor to decide how much importance is given to which neighbor by \textit{Eqn.} \ref{attention-weight}. It processes the target node with the neighboring nodes via a series of linear transformations.
\begin{equation}
    C_{j,k} = [\Theta(X_{p_{j_{node}}}), \Theta(X_{p_{k_{node}}}), \Theta_e(X_{p_{j_{edge}}}[k])].
\end{equation}
\begin{equation}
    \alpha_{j,k} = \frac{\exp(a(W^T(C_{j,k})))}{\sum_{p_k \in N(p_j)} \exp(a(W^T(C_{j,k})))}.
    \label{attention-weight}
\end{equation}

Finally node feature \(X_{p_{j_{node}}}\) is updated using \textit{Eqn. \ref{node-gat}}
\begin{equation}
    X_{p_{j_{node}}} = \alpha_{j,j}*\Theta(f_{p_j}) + \sum_{p_k\in N(p_j)}\alpha_{j,k}*\Theta(f_{p_k}).
    \label{node-gat}
\end{equation}
Here \(W\), \(\Theta\), and \(\Theta_e\) are learnable weights (actually FCN layer), \(a\) is the Leaky ReLU \cite{leaky-relu} activation function, and \([,]\) is concatenation operator. From \textit{Eqn. \ref{node-gat}} we get the processed node features for the target node \(p_j\). The process is repeated for all the nodes in \(V\). We use a dropout \cite{dropout} layer with the ratio of 0.5 inside the GAT layer. We use multiple consecutive GAT layers to process node features each one followed by a ReLU activation function except the last layer.

\subsubsection{Feature Aggregation}
At this point, we have multiple node level processed features. We have to aggregate them to find a graph level representation vector using an aggregation function. We tried a few options and found that average pooling works the best for the aggregation using \textit{Eqn. \ref{aggregation}}
\begin{equation}
    m_i = \frac{1}{n_i} \sum_{j=1}^{n_i} X_{p_{j_{node}}}.
    \label{aggregation}
\end{equation}

While processing data in a mini-batch of frames, we also need information about which points belong to which frame to calculate this aggregation, shown in Fig. \ref{frame-representation}.

\subsubsection{Frame Feature Processing Block (\(h_{frame}\))}
Frame feature processing block also consists of FCN layers each followed by ReLU activation, but it does not act like a shared-MLP because there is only one frame feature vector. The processed frame features \(X_{frame}\) are calculated as follows:
\begin{equation}
    X_{i_{frame}} = h_{frame}(f_i).
    \label{frame-feature}
\end{equation}

\begin{figure}[tp]
\centering\includegraphics[width=.45\textwidth]{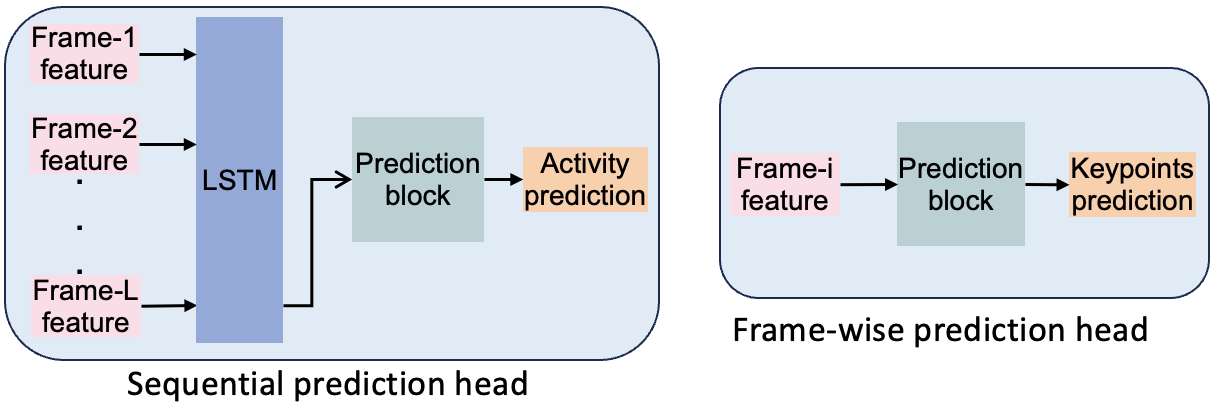}
\caption{Prediction heads.}
\label{prediction-heads}
\end{figure}

\subsubsection{Frame Feature Representation Vector}
Finally, we concatenate the aggregated feature vector \(X_{i_{frame}}\) with \(m_i\) to form the frame feature representation vector \(X_i\) which represents the entire point cloud for frame \(i\).

\subsubsection{Prediction Head}
We have two types of prediction head shown in Fig. \ref{prediction-heads}. For sequential modeling, we use Sequential prediction head. It takes \(L\) consecutive frames' representative feature vectors \(\{X_{(i-L+1)}, X_{(i+1-L+1)}, ..., X_i\}\), passes them through a bidirectional LSTM \cite{lstm} block, takes the last time-step's output, processes through prediction block \(h_{pred}\), and generates the final prediction. For frame-wise modeling, we do not need LSTM. Hence, we directly pass the frame representative feature vector \(X_i\) through \(h_{pred}\) to generate final prediction. Prediction block \(h_{pred}\) consists of FCN layers, each followed by a ReLU activation except the last layer. The final prediction from either frame-wise or sequential prediction head is denoted as \(X_{i_{pose}}\) for frame \(i\).

\section{Experimental Setup}
\subsection{Datasets}
\begin{table}[b]
\begin{center}
\caption{\textbf{Experimental Setup Details}\label{experimental-setup-details}}
\begin{adjustbox}{width=\columnwidth}
\begin{tabular}{||c|c|c|c|c||}

\hline
\textbf{Aspects} & \makecell{\textbf{MARS}\\frame-wise} & \makecell{\textbf{mRI}\\frame-wise} & \makecell{\textbf{MMFi}\\frame-wise} & \makecell{\textbf{MMActivity}\\sequential} \\ \hline \hline
Loss function & MSE & MPJPE & MPJPE & Cross-entropy\\
Evaluation metrics & RMSE, MAE & MPJPE, PA-MPJPE & MPJPE, PA-MPJPE & accuracy\\
Fusion frame & \(F=15\) & \(F=15\) & \(F=15\) & \(F=1\)\\
node downsampling & No & No & No & No\\
edge sampling by kNN & \(K=20\) & \(K=20\) & \(K=20\) & \(K=20\)\\
Input feature dimension (node, edge, frame) & (19, 6, 380) & (19, 6, 380) & (19, 6, 380) & (19, 6, 380)\\
Edge Processing Block FCN (layers, units) & (4, 64) & (4, 64) & (4, 64) & (5, 64)\\
Node Processing Block FCN (layers, units) & (4, 128) & (4, 128) & (4, 128) & (5, 128)\\
Frame Processing Block FCN (layers, units) & (4, 128) & (4, 128) & (4, 128) & (5, 128)\\
Graph Attention Block (layers, units, dropout) & (4, 128, 0.5) & (4, 128, 0.5) & (4, 128, 0.5) & (5, 128, 0.5)\\
LSTM block (layers, units, dropout) & - & - & - & (1, 64, 0.5)\\
Prediction Block FCN (layers, units) & (5, 128) & (5, 128) & (5, 128) & (5, 128)\\
Final output dimension & \(57(19\times3)\) & \(51(17\times3)\) & \(51(17\times3)\) & 5\\
Learning rate & 0.001 & 0.001 & 0.00075 & 0.0001\\
LR reduction factor & 0.995 & 0.975 & 0.99 & 0.99\\\hline

\end{tabular}
\end{adjustbox}
\end{center}
\end{table}
In this work we focus on efficient algorithm development to process point cloud data. Therefore, we test our approach on four publicly available mmWave datasets- MARS \cite{mars}, mRI \cite{mri}, MMFi \cite{mmfi} for HPE, and MMActivity \cite{radhar} for HAR. We describe several aspects such as metrics, loss function, hyper-parameters etc. in TABLE \ref{experimental-setup-details} for these datasets. We choose to use frame-wise HPE prediction for MARS, mRI, and MMFi datasets to keep consistency in the result comparison, as the original papers of these datasets used frame-wise HPE with frame fusion. For MMActivity dataset, we use sequential modeling similar to \cite{radhar} and other approaches for HAR.

Note that, for experiments on these datasets, we do not use node downsampling as the point clouds are not very dense. Yet, for scalability comparison, we discuss the trade-off between performance and time-resource complexity using node downsampling in the Ablation Study section. In the same section, we discuss the generalization of our approach on 3D object point cloud data using the ShapeNet dataset \cite{shapenet}. We use the node downsampling method on ShapeNet as the point clouds in ShapeNet are quite dense.

\subsection{Loss Function}
For the mRI and MMFi datasets, we train our model using Mean Per Joint Position Error (MPJPE) loss function after adjusting the skeleton position by mid-hip point using \textit{Eqn. \ref{loss-equation}}.
\begin{equation}
    X_{i_{adj_j}} = X_{i_{pose_j}} - X_{i_{pose_{mid-hip}}} + X_{i_{GT_{mid-hip}}}.
    \label{pose-adjustment}
\end{equation}
\begin{equation}
    \mathcal{L}_{MPJPE} = \frac{1}{B}{\sum_{i=1}^{B}\frac{1}{M}{\sum_{j=1}^{M}{||X_{i_{adj_j}} - X_{i_{GT_j}}||}}}.
    \label{loss-equation}
\end{equation}
Here, \(B\) is the batch-size for training, \(M\) is the number of keypoints in the skeleton, \(X_{i_{pose}}\) and \(X_{i_{GT}}\) indicate predicted pose and ground truth pose keypoints for frame \(i\) respectively. \(X_{i_{pose_j}}\) is the \(j\)th predicted keypoint for frame \(i\), \(X_{adj}\) denotes adjusted predicted keypoints, \(X_{i_{pose_{mid-hip}}}\) is the mid-hip keypoint for frame \(i\).

For the MARS dataset \cite{mars}, we use Mean Squared Error (MSE) loss for HPE. For the MMActivity\cite{radhar} dataset, we use cross-entropy loss for HAR.

\subsection{Hyper-parameters and Hardware Setup}
We show most of the model hyper-parameters in TABLE \ref{experimental-setup-details}. We conducted the whole experiment in Python \cite{python}. For deep learning model development, we used Pytorch \cite{pytorch}, for graph attention layer and model batch formation, we used Pytorch-Geometric \cite{pyg}. We conducted this experiment on a system with Intel i7-12700K 64-bit processor, 32 GB RAM, an Nvidia RTX-3090-ti GPU.
We set a large number of epochs for training, store the best performing model on validation set after each epoch and halt the training if no further convergence is observed for several epochs.
After finishing the training, we use the best model to infer on validation/test dataset for result collection.
\section{Result Analysis}
We discuss results on four datasets: MARS \cite{mars}, mRI \cite{mri}, MMFi \cite{mmfi} for HPE, and MMActivity \cite{radhar} for HAR. For HPE evaluation, we use metrics such as MPJPE, PA-MPJPE \cite{pa-mpjpe}, RMSE, and MAE, where lower scores are better. In this study, we emphasize on frame-wise modeling over sequential approach as more existing works are available in frame-wise approach to compare. For MPJPE calculation, we follow \textit{Eqn. \ref{loss-equation}}, as both \cite{mmfi} and \cite{mri} mentioned of using MPJPE of the adjusted skeleton by Thorax (mid-hip). We contacted the authors of \cite{mmfi} and \cite{mri} for confirmation of this adjustment, we received no response. However, the shared code from \cite{mmfi} suggests the MPJPE is calculated without skeleton adjustment, which may impact direct comparison on MPJPE. Nevertheless, PA-MPJPE remains unaffected by this adjustment, making it a more reliable metric for HPE. For HAR evaluation, we use accuracy, where higher scores are better.
We conducted extensive trainings across all dataset splits repeatedly, consistently observing stable results to validate the reproducibility and reliability. Finally, we recorded the results for those five datasets for each split set and reported here

\subsection{MARS \cite{mars} for HPE}
\begin{table}[b]
\begin{center}
\caption{\textbf{Result Comparison on MARS Dataset for HPE}\label{mars-results}}
\begin{adjustbox}{width=.98\columnwidth}
\begin{tabular}{||c|c|c|c|c|c||}

\hline
\textbf{approach} & \textbf{Model} & \makecell{\textbf{Average MAE}\\(cm)} & \makecell{\textbf{Average RMSE}\\(cm)} & \makecell{\textbf{MPJPE}\\(cm)} & \makecell{\textbf{PA-MPJPE}\\(cm)} \\ \hline \hline
\multirow{3}{*}{Frame-wise} & MARS \cite{mars} & 5.87 & 8.10 & - & - \\
 & Fast-Scalable \cite{fast-scalable} & 3.6 & - & - & - \\
 & mmGAT \cite{mmgat-original} & 3.42 & 5.653 & - & - \\
 & PCFEx (ours) & \textbf{2.454} & \textbf{4.330} & - & - \\\hline
\multirow{2}{*}{Sequential} & Multi-Frame \cite{shi} & 1.91 & 3.09 & - & - \\
 & Spatial-Temporal \cite{constrained-mmwave-hpe} & - & - & 3.05 & \textbf{1.97} \\
 & PCFEx-seq (ours) & \textbf{1.48} & \textbf{2.49} & \textbf{3.02} & 2.44\\ \hline

\end{tabular}
\end{adjustbox}
\end{center}
\end{table}
Originally, the MARS dataset was split into 60\% for training, 20\% for validation, and 20\% for testing. We follow the same splits, use RMSE, MAE metrics for frame-wise and MPJPE, PA-MPJPE for sequential approach, ensuring a fair comparison. In this work, our focus is more on frame-wise prediction where we achieve significantly lower RMSE and MAE errors than SOTA methods. The results on the test dataset are shown in TABLE \ref{mars-results} in centimeters (cm). To compare with \cite{constrained-mmwave-hpe} and \cite{shi}, we incorporate an LSTM block to develop a sequential HPE model, as shown in Fig. \ref{prediction-heads}. To prevent overfitting, we reduce the model's architecture, using 64 units and 3 layers for all blocks (node-processing, GAT, LSTM, prediction-block), except the LSTM block, which has 1 layer. We use frame fusion with \(F=3\) and 16 frames window for sequential input to LSTM same as \cite{constrained-mmwave-hpe}. Altogether, we see great reduction in MAE and RMSE over the previous models outperforming most of them in most metrics for both frame-wise and sequential modeling.

\subsection{mRI \cite{mri} for HPE}
The mRI dataset is divided into two settings: S1, a random 80/20 train-validation split, and S2, a subject-wise split with 16 subjects for training and 4 (subject 11, 13, 15, 17) for validation. The dataset contains 12 activities, with Protocol-1 (P1) including all activities and Protocol-2 (P2) consisting of 10 rehabilitation activities leading to four splits- S1P1, S1P2, S2P1, and S2P2. For evaluation, we use MPJPE and PA-MPJPE metrics in millimeter (mm) unit as in \cite{mri}, with MPJPE calculated using \textit{Eqn. \ref{loss-equation}}. The results are shown in TABLE \ref{mri-results}. We could improve HPE with 29\% and 25\% error reductions in MPJPE and PA-MPJPE scores, respectively over the SOTA methods. We see that the S1 splits show greater error reduction than the S2. Since \cite{mmskeleton} does not specify the split protocol, we omit their comparison on the mRI dataset. S2P1 in mRI and S2P3 in MMFi are analogous as both are subject-wise split containing all activities. MMFi's S2P3 MPJPE error is 72.74 mm, much lower than mRI S2P1's MPJPE error of 99.83. mRI dataset's S2 splits (subject-wise) show higher MPJPE error than MMFi dataset's S2 splits. On the other hand, S1 splits (random) show lower error on mRI dataset compared to MMFi dataset's S1 splits.
\begin{table}[t]
\begin{center}
\caption{\textbf{Result Comparison on mRI Dataset for HPE}\label{mri-results}}
\begin{adjustbox}{width=.9\columnwidth}
\begin{tabular}{||c|cc|cc|cc||}

\hline
\multirow{3}{*}{\textbf{Split mode}} & \multicolumn{2}{c|}{\textbf{mRI} \cite{mri}} & \multicolumn{2}{c|}{\textbf{mmGAT} \cite{mmgat-original}} & \multicolumn{2}{c||}{\textbf{PCFEx (ours)}} \\
  & \makecell{\textbf{MPJPE}\\(mm)}  & \makecell{\textbf{PA-MPJPE}\\(mm)}  & \makecell{\textbf{MPJPE}\\(mm)}  & \makecell{\textbf{PA-MPJPE}\\(mm)}  & \makecell{\textbf{MPJPE}\\(mm)} & \makecell{\textbf{PA-MPJPE}\\(mm)}  \\
\hline
\hline
S1P1 & 163.3 & 94.1 & 81 & 69 & \textbf{52.7} & \textbf{48.66} \\
S1P2 & 125.1 & 74.1 & 68 & 47 & \textbf{38.26} & \textbf{31.04} \\
S2P1 & 186.6 & 97.3 & 122 & 96 & \textbf{99.83} & \textbf{78.71} \\
S2P2 & 126.6 & 75 & 107 & 74 & \textbf{83.99} & \textbf{59.61} \\ \hline
\end{tabular}
\end{adjustbox}
\end{center}
\end{table}

\subsection{MMFi \cite{mmfi} for HPE}
\begin{table}[b]
\begin{center}
\caption{\textbf{Result Comparison on MMFi Dataset for HPE}\label{mmfi-results}}
\begin{adjustbox}{width=1\columnwidth}
\begin{tabular}{||c|c|cc|cc||}

\hline
\multirow{2}{*}{\textbf{Split mode}} & \makecell{\textbf{mmSkeleton}\cite{mmskeleton}\\(sequential)} & \multicolumn{2}{c|}{\makecell{\textbf{MM-Fi}\cite{mmfi}\\(frame-wise)}} & \multicolumn{2}{c||}{\makecell{\textbf{PCFEx (ours)}\\(frame-wise)}} \\
  & \makecell{\textbf{MPJPE}\\(mm)} & \makecell{\textbf{MPJPE}\\(mm)} & \makecell{\textbf{PA-MPJPE}\\(mm)} & \makecell{\textbf{MPJPE}\\(mm)} & \makecell{\textbf{PA-MPJPE}\\(mm)} \\
\hline
\hline
S1P1 & 88.85 & 109.8 & 55.6 & \textbf{37.36} & \textbf{32.74} \\
S1P2 & 74.56 & 124.3 & 55.3 & \textbf{36.0} & \textbf{31.75} \\
S1P3 & 71.02 & 117.0 & 57.3 & \textbf{45.17} & \textbf{38.57} \\
S2P1 & 102.5 & 128.4 & 58.7 & \textbf{70.63} & \textbf{53.83} \\
S2P2 & 117   & 142.2 & 57.4 & \textbf{68.61} & \textbf{54.54} \\
S2P3 & 108   & 129.7 & 60.0 & \textbf{72.74} & \textbf{57.09} \\
S3P1 &  -    & 166.2 & 73.9 & \textbf{87.84} & \textbf{67.55} \\
S3P2 &  -    & 168.0 & \textbf{73.0} & \textbf{93.81} & 76.1 \\
S3P3 &  -    & 161.6 & 73.7 & \textbf{90.79} & \textbf{72.94} \\ \hline
\end{tabular}
\end{adjustbox}
\end{center}
\end{table}
This dataset has three settings which are- random (S1), subject-wise (S2), and environment-wise (S3). They also have three protocols: P1 for 14 daily activities, P2 for 13 rehabilitation activities, and P3 for all 27 activities. In total, there are 9 different ways to split to train and validation sets from S1P1 to S3P3. We use MPJPE and PA-MPJPE scores same as \cite{mmfi} used to evaluate the results in millimeter (mm) unit. We use \textit{Eqn. \ref{loss-equation}} to calculate MPJPE. The comparison results are shown in \ref{mmfi-results}. MMFi is the largest and most diversified dataset for mmWave based HPE. Its random split (S1) has higher MPJPE error than mRI S1 split. S3 split (environment-wise) is the most difficult split to train the model. It indicates that generalization of pose estimation is difficult across different environment. Although we were able to make significant improvements to reduce the PA-MPJPE scores over \cite{mmfi} reported scores, we believe that more improvements are necessary for the real-world applications.

\subsection{MMActivity \cite{radhar} for HAR}
\begin{table}[t]
\begin{center}
\caption{\textbf{Result Comparison on MMActivity Dataset for HAR} \label{radhar-results}}
\begin{adjustbox}{width=.5\columnwidth}
\begin{tabular}{||c|c||}

\hline
\textbf{Model} & \textbf{Accuracy (\%)} \\ \hline \hline
Radhar \cite{radhar} & 90.47 \\
Tesla-Rapture \cite{tesla-rapture} & 96.97 \\
mmPoint-GNN \cite{mmpoint-gnn} & 96.97 \\
Fast HAR \cite{fast-har} & 96.99 \\
Noninvasive \cite{noninvasive-har}& 97.8 \\
PCFEx (ours) & \textbf{98.8} \\
\hline
\end{tabular}
\end{adjustbox}
\end{center}
\end{table}
MMActivity dataset is introduced in \cite{radhar}. They divided the dataset into train and test and publicly shared it. We used the same split with a similar setup of sequence length 60 frames with 10 frames interval as mentioned in \cite{radhar}. The result comparison is shown in TABLE \ref{radhar-results}.
During our model training, we found that frame features improved the training significantly. Frame features allow a shortcut path to pass the data from the input point cloud to the frame feature representation vector. It results in good improvement of the model training when subtle information are not essential.

\subsection{Model complexity analysis}
To enable advanced feature processing, which requires substantial preparation time, we compare feature extraction time, training time, model complexity, and evaluation performance with two other benchmarks. We use the S2P2 setting from the MMFi \cite{mmfi} dataset, used their officially released script for this comparison, and adapt the model from Fast-scalable \cite{fast-scalable} with slight modification for MMFi dataset. The comparison results are shown in Table \ref{complexity-comparison-table}. Despite having fewer trainable parameters, our approach takes significantly longer for feature processing and requires more training time than the other two models. However, our approach achieves great performance improvement over the other two benchmarks.
\begin{table}[tb]
\begin{center}
\caption{\textbf{Comparison on model complexity and runtime on S2P2 setting from MMFi \cite{mmfi} dataset.} \label{complexity-comparison-table}}
\begin{adjustbox}{width=\columnwidth}
\begin{tabular}{||c|c|c|c|c|c||}

\hline
\textbf{Model} & \textbf{Feature processing time} & \makecell{\textbf{Training time}\\(100 epoch)} & \textbf{Trainable parameters} & \makecell{\textbf{MPJPE}\\(mm)} & \makecell{\textbf{PA-MPJPE}\\(mm)} \\ \hline \hline

Fast-scalable \cite{fast-scalable} & 2m 45s & 29m 10s & 303,923 & 88 & 73 \\
MMFi \cite{mmfi} & 1m, 32s & 3h, 12m & 1,454,515 & 73.1 & 57 \\
PCFEx (ours) & 1h, 25m & 3h, 44m & 352,691 & 67.7 & 53.7 \\\hline

\end{tabular}
\end{adjustbox}
\end{center}
\end{table}
\section{Ablation Study}

\begin{table}[t]
\caption{\textbf{Contributions of individual building blocks}\label{progression}}
\begin{adjustbox}{width=1.02\columnwidth}
\begin{tabular}{||c|c|cc|cc|cc|c||}

\hline
\multirow{3}{*}{\raisebox{-1.5\totalheight}{\rotatebox[]{90}{\textbf{Modes}}}} & \multirow{3}{*}{\raisebox{-6\totalheight}{\textbf{Combinations}}} & \multicolumn{6}{c|}{\textbf{HPE}} & \textbf{HAR} \\
\cline{3-9}
 &  & \multicolumn{2}{c|}{\textbf{MMFi-S2P1}} & \multicolumn{2}{c|}{\textbf{MMFi-S2P2}} & \multicolumn{2}{c|}{\textbf{MARS}} & \makecell{\textbf{MM}\\\textbf{Activity}} \\
 \cline{3-9}
 &  & \makecell{MPJPE\\(mm)} & \makecell{PA-\\MPJPE\\(mm)} & \makecell{MPJPE\\(mm)} & \makecell{PA-\\MPJPE\\(mm)} & \makecell{MAE\\(cm)} & \makecell{RMSE\\(cm)} & Accuracy\\
\hline
1 & Basic node (5D) & 75.63 & 58.44 & 76.16 & 61.79 & 2.715 & 4.642 & 95.01 \\
2 & Basic node + frame feature & 76.63 & 58.79 & 75.70 & 61.73 & 2.700 & 4.594 & 97.72 \\
3 & Node feature (5D + 14D) & 71.05 & 54.18 & 68.96 & 55.04 & \textbf{2.427} & \textbf{4.257} & 94.47 \\
4 & Node + Frame feature & 72.07 & 54.67 & 69.05 & 54.90 & 2.574 & 4.540 & 98.06 \\
5 & Node+Edge+Frame feature & \textbf{70.63} & \textbf{53.83} & \textbf{68.61} & \textbf{54.54} & 2.454 & 4.330 & \textbf{98.80} \\ \hline
\end{tabular}
\end{adjustbox}
\end{table}
\subsection{Contributions of individual building blocks}
To discuss the individual contributions of different building blocks of our model, We have conducted experimental studies in 5 modes shown in TABLE \ref{progression}. We used two splits (S2P1 and S2P2) from MMFi dataset \cite{mmfi}, whole MARS dataset \cite{mars} for HPE and MMActivity dataset \cite{radhar} for HAR in this study. Our obtained results in TABLE \ref{progression} suggest that node feature extraction gives large performance boost in our method in HPE tasks. We have also noticed that frame features significantly help the model in HAR tasks. From these experiments on HPE, we see a trend that node feature extraction (mode-3) drastically reduces the MPJPE and PA-MPJPE errors over the basic node features (mode-1). Frame features’ impact is less visible on HPE tasks. Combining node, edge and frame features perform the best in most cases for HPE. For HAR, frame features play the key role in enhancing prediction accuracy. Frame features significantly improve accuracy over basic node feature. The node feature extraction shows less impact in mode-3 over mode-1, mode-2 which indicates that HAR requires less details of the point-cloud. Finally, the combination of node, edge and frame features performs the best for HAR.

Our assumption is that HPE requires more detailed information of the point-cloud to accurately predict the pose keypoints where HAR requires less detailed information regarding the point-cloud. Frame features allow the model a shorter propagation path (analogous to skip-connections) and helps the model in tasks that require less exploration of point-cloud itself. Hence, we see that frame features significantly improve the performance in HAR task in MMActivity dataset where the impact of frame features is less visible in HPE.

\begin{table}[b]
\begin{center}
\caption{\textbf{Classification Accuracy on ShapeNet Dataset} \label{shapenet-results}}
\begin{adjustbox}{width=.8\columnwidth}
\begin{tabular}{||c|c||}

\hline
\textbf{Experiment} & \textbf{Accuracy}\\ \hline
Without node \& frame feature extraction & 74.57 \\
with node \& frame feature extraction & 78.18 \\
\hline
\end{tabular}
\end{adjustbox}
\end{center}
\end{table}
\subsection{Generalized impact of node feature extraction}
Our node feature extraction can be integrated to any point-cloud modeling approach as a plug-and-play module irrespective of the foundational model architecture such as CNN, Transformer, MLP or GNN-based approach as our technique only extends the feature dimension of input node features with significant performance enhancements.

We believe that feature extraction process at the node, edge, and frame levels is a generalizable method, applicable to not only mmWave but also other point cloud data. To test this, we applied our approach to the ShapeNet dataset \cite{shapenet}, which includes 3D point cloud data for 55 real-world objects. We split the dataset into 80/20 train-validation ratio for classification. Our results in TABLE \ref{shapenet-results} show that node and frame feature extraction significantly improves classification accuracy. Given ShapeNet’s dense point clouds, we employed node downsampling with a cell width of (0.01, 0.01, 0.01). We trained the model from scratch, excluding multi-view techniques, vertex normals, surface point cloud sampling, and data augmentation, as these are outside the scope of this study. As a result, our classification accuracy is lower than state-of-the-art results (max 85.9\% \cite{point-xformer-nico}).

\begin{table}[t]
\begin{center}
\caption{\textbf{Complexity-Performance Trade-off by Node-sampling} \label{complexity-performance}}
\begin{adjustbox}{width=.99\columnwidth}
\begin{tabular}{||c|c|c|c|c||}

\hline
\textbf{Dataset} & \textbf{Observation point} & \makecell{\textbf{Downsampling}\\(Q=1)} & \makecell{\textbf{Downsampling}\\(Q=5)} & \textbf{\makecell{Without node \\downsampling}} \\ \hline
\multirow{6}{*}{MMFi-S2P1} & Data processing time & 36m & 44m & 1h 46m \\
 & Feature data storage memory & 23.9 GB & 69.6 GB & 110 GB\\
 & Model training time & 2h 5m & 4h 6m & 5h 50m \\
 & MPJPE (mm) & 76.83 & 72.32 & 70.9 \\
 & PA-MPJPE (mm) & 60.2 & 55.93 & 53.64  \\
 & \makecell{average GPU usage\\(Batch=32)} & 3100 MB & 13300 MB & 23900 MB\\
\hline
\end{tabular}
\end{adjustbox}
\end{center}
\end{table}
\subsection{Complexity vs performance trade-off}
In TABLE \ref{complexity-performance}, we highlight the effectiveness of node-sampling when resource allocation and data processing time are crucial. Node-sampling allows training a reliable model with limited resources, conceding minimal performance degradation.
Because of 3.75cm resolution in mmWave radars, we apply node downsampling with a cell width of (0.035m, 0.035m, 0.035m) along \(x, y, z\) axes for \(Q=1\), \(Q=5\) and no downsampling, three options all trained for 100 epochs. This process mostly limits the redundant points in the point-cloud. The number of edges drop rapidly which contributes the most in reducing execution time and storage complexity. StatBox backed node and frame feature extraction blocks are affected the most resulting in slight reduction in HPE accuracy.

\subsection{Other observations}
We have two additional observations: First, fewer frame fusions tend to work better with sequential modeling. When we trained the model for the MMActivity dataset \cite{radhar}, we first noticed this occurrence. We also noticed similar occurrence in HPE training while experimenting sequential modeling for the MARS \cite{mars} dataset. Second, while using frame features in sequential HPE models, we observed slight MPJPE degradation, but we see significant improvement in HAR accuracy.
\section{Conclusion}
We propose a graph neural network model for point cloud data, focusing on advanced feature extraction and processing techniques. Our approach shows strong potential along with a few limitations. Compared to existing feature extraction models, our approach offers a flexible, plug-and-play Statbox module that enables semantically meaningful and task-specific feature design—enhancing scalability across diverse tasks (HPE, HAR, point-cloud classification) and varying point-clouds, while reducing susceptibility to background noise and outliers. It currently supports HPE and HAR for a single person per frame. Again, all datasets were collected in controlled environments, with only the MMFi dataset \cite{mmfi} providing data across multiple environments, which proved to be the hardest to achieve lower MPJPE scores for HPE. Despite these challenges, we demonstrate the effectiveness of our method on four benchmark datasets, surpassing the previous benchmarks with significant margins. In future works, we aim to address these limitations by incorporating real-world mmWave data, enhancing point cloud processing, and expanding our model to support multi-person HPE and HAR.

\section{Acknowledgement}
This work was supported by JST ASPIRE Grant Number JPMJAP2326, Japan.

\bibliographystyle{IEEEtran}
\bibliography{ref.bib}

@article{mars,
author = {An, Sizhe and Ogras, Umit Y.},
title = {MARS: mmWave-based Assistive Rehabilitation System for Smart Healthcare},
year = {2021},
issue_date = {October 2021},
publisher = {Association for Computing Machinery},
address = {New York, NY, USA},
volume = {20},
number = {5s},
issn = {1539-9087},
url = {https://doi.org/10.1145/3477003},
doi = {10.1145/3477003},
journal = {ACM Trans. Embed. Comput. Syst.},
month = {sep},
articleno = {72},
numpages = {22}
}

@inproceedings{mri,
 author = {An, Sizhe and Li, Yin and Ogras, Umit},
 booktitle = {Advances in Neural Information Processing Systems},
 editor = {S. Koyejo and S. Mohamed and A. Agarwal and D. Belgrave and K. Cho and A. Oh},
 pages = {27414--27426},
 publisher = {Curran Associates, Inc.},
 title = {mRI: Multi-modal 3D Human Pose Estimation Dataset using mmWave, RGB-D, and Inertial Sensors},
 volume = {35},
 year = {2022}
}

@inproceedings{radhar,
author = {Singh, Akash Deep and Sandha, Sandeep Singh and Garcia, Luis and Srivastava, Mani},
title = {RadHAR: Human Activity Recognition from Point Clouds Generated through a Millimeter-Wave Radar},
year = {2019},
isbn = {9781450369329},
publisher = {Association for Computing Machinery},
address = {New York, NY, USA},
url = {https://doi.org/10.1145/3349624.3356768},
doi = {10.1145/3349624.3356768},
booktitle = {Proceedings of the 3rd ACM Workshop on Millimeter-Wave Networks and Sensing Systems},
pages = {51–56},
numpages = {6},
location = {Los Cabos, Mexico},
series = {mmNets'19}
}

@inproceedings{fast-scalable,
author = {An, Sizhe and Ogras, Umit Y.},
title = {Fast and Scalable Human Pose Estimation Using MmWave Point Cloud},
year = {2022},
isbn = {9781450391429},
publisher = {Association for Computing Machinery},
address = {New York, NY, USA},
url = {https://doi.org/10.1145/3489517.3530522},
doi = {10.1145/3489517.3530522},
booktitle = {Proceedings of the 59th ACM/IEEE Design Automation Conference},
pages = {889–894},
numpages = {6},
location = {San Francisco, California},
series = {DAC '22}
}

@INPROCEEDINGS{hpe-part-affinity,
  author={Cao, Zhe and Simon, Tomas and Wei, Shih-En and Sheikh, Yaser},
  booktitle={2017 IEEE Conference on Computer Vision and Pattern Recognition (CVPR)}, 
  title={Realtime Multi-person 2D Pose Estimation Using Part Affinity Fields}, 
  year={2017},
  volume={},
  number={},
  pages={1302-1310},
  doi={10.1109/CVPR.2017.143}
}

@article{openpose,
author = {Cao, Zhe and Hidalgo, Gines and Simon, Tomas and Wei, Shih-En and Sheikh, Yaser},
title = {OpenPose: Realtime Multi-Person 2D Pose Estimation Using Part Affinity Fields},
year = {2021},
issue_date = {Jan. 2021},
publisher = {IEEE Computer Society},
address = {USA},
volume = {43},
number = {1},
issn = {0162-8828},
url = {https://doi.org/10.1109/TPAMI.2019.2929257},
doi = {10.1109/TPAMI.2019.2929257},
journal = {IEEE Trans. Pattern Anal. Mach. Intell.},
month = {jan},
pages = {172–186},
numpages = {15}
}

@ARTICLE {hrnet,
author = {J. Wang and K. Sun and T. Cheng and B. Jiang and C. Deng and Y. Zhao and D. Liu and Y. Mu and M. Tan and X. Wang and W. Liu and B. Xiao},
journal = {IEEE Transactions on Pattern Analysis \& Machine Intelligence},
title = {Deep High-Resolution Representation Learning for Visual Recognition},
year = {2021},
volume = {43},
number = {10},
issn = {1939-3539},
pages = {3349-3364},
doi = {10.1109/TPAMI.2020.2983686},
publisher = {IEEE Computer Society},
address = {Los Alamitos, CA, USA},
month = {oct}
}

@INPROCEEDINGS {higher-hrnet,
author = {B. Cheng and B. Xiao and J. Wang and H. Shi and T. S. Huang and L. Zhang},
booktitle = {2020 IEEE/CVF Conference on Computer Vision and Pattern Recognition (CVPR)},
title = {HigherHRNet: Scale-Aware Representation Learning for Bottom-Up Human Pose Estimation},
year = {2020},
volume = {},
issn = {},
pages = {5385-5394},
doi = {10.1109/CVPR42600.2020.00543},
url = {https://doi.ieeecomputersociety.org/10.1109/CVPR42600.2020.00543},
publisher = {IEEE Computer Society},
address = {Los Alamitos, CA, USA},
month = {jun}
}

@inproceedings{one-stage-1,
author = {Sun, Xiao and Xiao, Bin and Wei, Fangyin and Liang, Shuang and Wei, Yichen},
title = {Integral Human Pose Regression},
year = {2018},
isbn = {978-3-030-01230-4},
publisher = {Springer-Verlag},
address = {Berlin, Heidelberg},
url = {https://doi.org/10.1007/978-3-030-01231-1_33},
doi = {10.1007/978-3-030-01231-1_33},
booktitle = {Computer Vision – ECCV 2018: 15th European Conference, Munich, Germany, September 8–14, 2018, Proceedings, Part VI},
pages = {536–553},
numpages = {18},
location = {Munich, Germany}
}

@article{one-stage-2,
  title={2D/3D Pose Estimation and Action Recognition Using Multitask Deep Learning},
  author={Diogo Carbonera Luvizon and David Picard and Hedi Tabia},
  journal={2018 IEEE/CVF Conference on Computer Vision and Pattern Recognition},
  year={2018},
  pages={5137-5146},
  url={https://api.semanticscholar.org/CorpusID:3621625}
}

@misc{one-stage-3,
author = {Chun, Sungho and Park, Sungbum and Chang, Ju},
year = {2023},
month = {06},
pages = {},
title = {Representation learning of vertex heatmaps for 3D human mesh reconstruction from multi-view images}
}

@misc{one-stage-4,
      title={Learnable human mesh triangulation for 3D human pose and shape estimation}, 
      author={Sungho Chun and Sungbum Park and Ju Yong Chang},
      year={2022},
      eprint={2208.11251},
      archivePrefix={arXiv},
      primaryClass={cs.CV}
}

@misc{two-stage-1,
      title={MotionBERT: A Unified Perspective on Learning Human Motion Representations}, 
      author={Wentao Zhu and Xiaoxuan Ma and Zhaoyang Liu and Libin Liu and Wayne Wu and Yizhou Wang},
      year={2023},
      eprint={2210.06551},
      archivePrefix={arXiv},
      primaryClass={cs.CV}
}

@misc{two-stage-2,
      title={3D human pose estimation in video with temporal convolutions and semi-supervised training}, 
      author={Dario Pavllo and Christoph Feichtenhofer and David Grangier and Michael Auli},
      year={2019},
      eprint={1811.11742},
      archivePrefix={arXiv},
      primaryClass={cs.CV}
}

@inbook{two-stage-3,
   title={Exploiting Temporal Information for 3D Human Pose Estimation},
   ISBN={9783030012496},
   ISSN={1611-3349},
   url={http://dx.doi.org/10.1007/978-3-030-01249-6_5},
   DOI={10.1007/978-3-030-01249-6_5},
   booktitle={Lecture Notes in Computer Science},
   publisher={Springer International Publishing},
   author={Hossain, Mir Rayat Imtiaz and Little, James J.},
   year={2018},
   pages={69–86} }

@inproceedings{rfpose3d,
author = {Zhao, Mingmin and Tian, Yonglong and Zhao, Hang and Alsheikh, Mohammad Abu and Li, Tianhong and Hristov, Rumen and Kabelac, Zachary and Katabi, Dina and Torralba, Antonio},
title = {RF-based 3D skeletons},
year = {2018},
isbn = {9781450355674},
publisher = {Association for Computing Machinery},
address = {New York, NY, USA},
url = {https://doi.org/10.1145/3230543.3230579},
doi = {10.1145/3230543.3230579},
booktitle = {Proceedings of the 2018 Conference of the ACM Special Interest Group on Data Communication},
pages = {267–281},
numpages = {15},
location = {Budapest, Hungary},
series = {SIGCOMM '18}
}

@ARTICLE{mmpose,
  author={Sengupta, Arindam and Jin, Feng and Zhang, Renyuan and Cao, Siyang},
  journal={IEEE Sensors Journal}, 
  title={mm-Pose: Real-Time Human Skeletal Posture Estimation Using mmWave Radars and CNNs}, 
  year={2020},
  volume={20},
  number={17},
  pages={10032-10044},
  doi={10.1109/JSEN.2020.2991741}}

@INPROCEEDINGS{mmgat-original,
  author={Abdullah Al, Masud and Shi, Xintong and Mondher, Bouazizi and Ohtsuki, Tomoaki},
  booktitle={ICC 2024 - IEEE International Conference on Communications}, 
  title="{mmGAT: Pose Estimation by Graph Attention with Mutual Features from mmWave Radar Point Cloud}", 
  year={2024},
  volume={},
  number={},
  pages={2161-2166},
  doi={10.1109/ICC51166.2024.10622791}}

@INPROCEEDINGS{mmgat,
  author={Anonimous},
  booktitle={Included in supplimental materials as "our-previous-work"}}

@misc{mmfi,
      title={MM-Fi: Multi-Modal Non-Intrusive 4D Human Dataset for Versatile Wireless Sensing}, 
      author={Jianfei Yang and He Huang and Yunjiao Zhou and Xinyan Chen and Yuecong Xu and Shenghai Yuan and Han Zou and Chris Xiaoxuan Lu and Lihua Xie},
      year={2023},
      eprint={2305.10345},
      archivePrefix={arXiv},
      primaryClass={eess.SP},
      url={https://arxiv.org/abs/2305.10345}, 
}

@article{lstm,
    author = {Hochreiter, Sepp and Schmidhuber, Jürgen},
    title = "{Long Short-Term Memory}",
    journal = {Neural Computation},
    volume = {9},
    number = {8},
    pages = {1735-1780},
    year = {1997},
    month = {11},
    issn = {0899-7667},
    doi = {10.1162/neco.1997.9.8.1735},
    url = {https://doi.org/10.1162/neco.1997.9.8.1735},
    eprint = {https://direct.mit.edu/neco/article-pdf/9/8/1735/813796/neco.1997.9.8.1735.pdf},
}

@article{joint-global-local,
  title={A joint global--local network for human pose estimation with millimeter wave radar},
  author={Cao, Zhongping and Ding, Wen and Chen, Rihui and Zhang, Jianxiong and Guo, Xuemei and Wang, Guoli},
  journal={IEEE Internet of Things Journal},
  volume={10},
  number={1},
  pages={434--446},
  year={2022},
  publisher={IEEE}
}

@article{throughwall-xformer,
  title={A study on 3D human pose estimation using through-wall IR-UWB radar and transformer},
  author={Kim, Gon Woo and Lee, Sang Won and Son, Ha Young and Choi, Kae Won},
  journal={IEEE Access},
  volume={11},
  pages={15082--15095},
  year={2023},
  publisher={IEEE}
}

@article{mi-mesh,
  title={MI-Mesh: 3D human mesh construction by fusing image and millimeter wave},
  author={Ding, Han and Chen, Zhenbin and Zhao, Cui and Wang, Fei and Wang, Ge and Xi, Wei and Zhao, Jizhong},
  journal={Proceedings of the ACM on Interactive, Mobile, Wearable and Ubiquitous Technologies},
  volume={7},
  number={1},
  pages={1--24},
  year={2023},
  publisher={ACM New York, NY, USA}
}

@misc{hupr,
      title={HuPR: A Benchmark for Human Pose Estimation Using Millimeter Wave Radar}, 
      author={Shih-Po Lee and Niraj Prakash Kini and Wen-Hsiao Peng and Ching-Wen Ma and Jenq-Neng Hwang},
      year={2022},
      eprint={2210.12564},
      archivePrefix={arXiv},
      primaryClass={cs.CV},
      url={https://arxiv.org/abs/2210.12564}, 
}

@inproceedings{xformer,
 author = {Vaswani, Ashish and Shazeer, Noam and Parmar, Niki and Uszkoreit, Jakob and Jones, Llion and Gomez, Aidan N and Kaiser, \L ukasz and Polosukhin, Illia},
 booktitle = {Advances in Neural Information Processing Systems},
 editor = {I. Guyon and U. Von Luxburg and S. Bengio and H. Wallach and R. Fergus and S. Vishwanathan and R. Garnett},
 pages = {},
 publisher = {Curran Associates, Inc.},
 title = {Attention is All you Need},
 url = {https://proceedings.neurips.cc/paper_files/paper/2017/file/3f5ee243547dee91fbd053c1c4a845aa-Paper.pdf},
 volume = {30},
 year = {2017}
}

@misc{gat,
      title={Graph Attention Networks}, 
      author={Petar Veličković and Guillem Cucurull and Arantxa Casanova and Adriana Romero and Pietro Liò and Yoshua Bengio},
      year={2018},
      eprint={1710.10903},
      archivePrefix={arXiv},
      primaryClass={stat.ML},
      url={https://arxiv.org/abs/1710.10903}, 
}

@misc{pointnet++,
      title={PointNet++: Deep Hierarchical Feature Learning on Point Sets in a Metric Space}, 
      author={Charles R. Qi and Li Yi and Hao Su and Leonidas J. Guibas},
      year={2017},
      eprint={1706.02413},
      archivePrefix={arXiv},
      primaryClass={cs.CV},
      url={https://arxiv.org/abs/1706.02413}, 
}

@inproceedings{mm-mesh,
  title={mmMesh: Towards 3D real-time dynamic human mesh construction using millimeter-wave},
  author={Xue, Hongfei and Ju, Yan and Miao, Chenglin and Wang, Yijiang and Wang, Shiyang and Zhang, Aidong and Su, Lu},
  booktitle={Proceedings of the 19th Annual International Conference on Mobile Systems, Applications, and Services},
  pages={269--282},
  year={2021}
}

@article{joint-dynamic,
  title={Human parsing with joint learning for dynamic mmwave radar point cloud},
  author={Wang, Shuai and Cao, Dongjiang and Liu, Ruofeng and Jiang, Wenchao and Yao, Tianshun and Lu, Chris Xiaoxuan},
  journal={Proceedings of the ACM on Interactive, Mobile, Wearable and Ubiquitous Technologies},
  volume={7},
  number={1},
  pages={1--22},
  year={2023},
  publisher={ACM New York, NY, USA}
}

@misc{pointnet,
      title={PointNet: Deep Learning on Point Sets for 3D Classification and Segmentation}, 
      author={Charles R. Qi and Hao Su and Kaichun Mo and Leonidas J. Guibas},
      year={2017},
      eprint={1612.00593},
      archivePrefix={arXiv},
      primaryClass={cs.CV},
      url={https://arxiv.org/abs/1612.00593}, 
}

@inproceedings{fast-rfpose,
  title={Fast 3D Human Pose Estimation Using RF Signals},
  author={Yu, Cong and Zhang, Dongheng and Wu, Zhi and Xie, Chunyang and Lu, Zhi and Hu, Yang and Chen, Yan},
  booktitle={ICASSP 2023-2023 IEEE International Conference on Acoustics, Speech and Signal Processing (ICASSP)},
  pages={1--5},
  year={2023},
  organization={IEEE}
}

@inproceedings{point-xformer,
  title={Point transformer},
  author={Zhao, Hengshuang and Jiang, Li and Jia, Jiaya and Torr, Philip HS and Koltun, Vladlen},
  booktitle={Proceedings of the IEEE/CVF international conference on computer vision},
  pages={16259--16268},
  year={2021}
}

@inproceedings{xformer-hpe,
  title={A Transformer-Based Network for Human Pose Estimation using Millimeter Wave Radar Data},
  author={Wei, Guiyan and Cui, Chang and Dong, Xichao},
  booktitle={2023 International Applied Computational Electromagnetics Society Symposium (ACES-China)},
  pages={1--4},
  year={2023},
  organization={IEEE}
}

@inproceedings{relu,
  title={Rectified linear units improve restricted boltzmann machines},
  author={Nair, Vinod and Hinton, Geoffrey E},
  booktitle={Proceedings of the 27th international conference on machine learning (ICML-10)},
  pages={807--814},
  year={2010}
}

@article{dropout,
  title={Dropout: a simple way to prevent neural networks from overfitting},
  author={Srivastava, Nitish and Hinton, Geoffrey and Krizhevsky, Alex and Sutskever, Ilya and Salakhutdinov, Ruslan},
  journal={The journal of machine learning research},
  volume={15},
  number={1},
  pages={1929--1958},
  year={2014},
  publisher={JMLR. org}
}

@inproceedings{leaky-relu,
  title={Rectifier nonlinearities improve neural network acoustic models},
  author={Maas, Andrew L and Hannun, Awni Y and Ng, Andrew Y and others},
  booktitle={Proc. icml},
  volume={30},
  pages={3},
  year={2013},
  organization={Atlanta, GA}
}

@INPROCEEDINGS{shi,
  author={Shi, Xintong and Ohtsuki, Tomoaki},
  booktitle={2023 IEEE SENSORS}, 
  title={A Robust Multi-Frame mmWave Radar Point Cloud-Based Human Skeleton Estimation Approach with Point Cloud Reliability Assessment}, 
  year={2023},
  volume={},
  number={},
  pages={1-4},
  doi={10.1109/SENSORS56945.2023.10325204}}

@misc{pytorch,
      title={PyTorch: An Imperative Style, High-Performance Deep Learning Library}, 
      author={Adam Paszke and Sam Gross and Francisco Massa and Adam Lerer and James Bradbury and Gregory Chanan and Trevor Killeen and Zeming Lin and Natalia Gimelshein and Luca Antiga and Alban Desmaison and Andreas Köpf and Edward Yang and Zach DeVito and Martin Raison and Alykhan Tejani and Sasank Chilamkurthy and Benoit Steiner and Lu Fang and Junjie Bai and Soumith Chintala},
      year={2019},
      eprint={1912.01703},
      archivePrefix={arXiv},
      primaryClass={cs.LG},
      url={https://arxiv.org/abs/1912.01703}, 
}

@misc{pyg,
      title={Fast Graph Representation Learning with PyTorch Geometric}, 
      author={Matthias Fey and Jan Eric Lenssen},
      year={2019},
      eprint={1903.02428},
      archivePrefix={arXiv},
      primaryClass={cs.LG},
      url={https://arxiv.org/abs/1903.02428}, 
}

@book{python,
 author = {Van Rossum, Guido and Drake, Fred L.},
 title = {Python 3 Reference Manual},
 year = {2009},
 isbn = {1441412697},
 publisher = {CreateSpace},
 address = {Scotts Valley, CA}
}

@article{pa-mpjpe,
  title={Procrustes analysis},
  author={Ross, Amy},
  journal={Course report, Department of Computer Science and Engineering, University of South Carolina},
  volume={26},
  pages={1--8},
  year={2004},
  publisher={Citeseer}
}

@INPROCEEDINGS{mmpoint-gnn,
  author={Gong, Peixian and Wang, Chunyu and Zhang, Lihua},
  booktitle={2021 International Joint Conference on Neural Networks (IJCNN)}, 
  title={MMPoint-GNN: Graph Neural Network with Dynamic Edges for Human Activity Recognition through a Millimeter-Wave Radar}, 
  year={2021},
  volume={},
  number={},
  pages={1-7},
  doi={10.1109/IJCNN52387.2021.9533989}}

@Article{improving-har-gnn,
AUTHOR = {Lee, Gawon and Kim, Jihie},
TITLE = {Improving Human Activity Recognition for Sparse Radar Point Clouds: A Graph Neural Network Model with Pre-Trained 3D Human-Joint Coordinates},
JOURNAL = {Applied Sciences},
VOLUME = {12},
YEAR = {2022},
NUMBER = {4},
ARTICLE-NUMBER = {2168},
URL = {https://www.mdpi.com/2076-3417/12/4/2168},
ISSN = {2076-3417},
}

@misc{st-gcn,
      title={Spatial Temporal Graph Convolutional Networks for Skeleton-Based Action Recognition}, 
      author={Sijie Yan and Yuanjun Xiong and Dahua Lin},
      year={2018},
      eprint={1801.07455},
      archivePrefix={arXiv},
      primaryClass={cs.CV},
      url={https://arxiv.org/abs/1801.07455}, 
}

@inproceedings{fast-har,
author = {Shao, Tongfei and Du, Zheyu and Li, Chuanyou and Wu, Tianxing and Wang, Meng},
title = {Fast Human Action Recognition via Millimeter Wave Radar Point Cloud Sequences Learning},
year = {2024},
isbn = {9798400704369},
publisher = {Association for Computing Machinery},
address = {New York, NY, USA},
url = {https://doi.org/10.1145/3627673.3679787},
doi = {10.1145/3627673.3679787},
booktitle = {Proceedings of the 33rd ACM International Conference on Information and Knowledge Management},
pages = {2024–2033},
numpages = {10},
location = {Boise, ID, USA},
series = {CIKM '24}
}

@ARTICLE{noninvasive-har,
  author={Yu, Chengxi and Xu, Zhezhuang and Yan, Kun and Chien, Ying-Ren and Fang, Shih-Hau and Wu, Hsiao-Chun},
  journal={IEEE Systems Journal}, 
  title={Noninvasive Human Activity Recognition Using Millimeter-Wave Radar}, 
  year={2022},
  volume={16},
  number={2},
  pages={3036-3047},
  doi={10.1109/JSYST.2022.3140546}}

@ARTICLE{constrained-mmwave-hpe,
  author={Chen, Lin and Guo, Xuemei and Wang, Guoli and Li, Hongyi},
  journal={IEEE Transactions on Cognitive and Developmental Systems}, 
  title={Spatial–Temporal Multiscale Constrained Learning for mmWave-Based Human Pose Estimation}, 
  year={2024},
  volume={16},
  number={3},
  pages={1108-1120},
  doi={10.1109/TCDS.2023.3334302}}

@INPROCEEDINGS{mmskeleton,
  author={Li, Wei and Lei, Wen and Shi, Kun and Shi, Zhiguo and Wang, Yong and Zhou, Jinhai},
  booktitle={2024 IEEE/CIC International Conference on Communications in China (ICCC)}, 
  title={mmSkeleton: 3D Human Skeleton Estimation Using Millimeter Wave Radar Sparse Point Clouds}, 
  year={2024},
  volume={},
  number={},
  pages={307-312},
  doi={10.1109/ICCC62479.2024.10681946}}

@misc{shapenet,
      title={ShapeNet: An Information-Rich 3D Model Repository}, 
      author={Angel X. Chang and Thomas Funkhouser and Leonidas Guibas and Pat Hanrahan and Qixing Huang and Zimo Li and Silvio Savarese and Manolis Savva and Shuran Song and Hao Su and Jianxiong Xiao and Li Yi and Fisher Yu},
      year={2015},
      eprint={1512.03012},
      archivePrefix={arXiv},
      primaryClass={cs.GR},
      url={https://arxiv.org/abs/1512.03012}, 
}

@misc{tesla-rapture,
      title={Tesla-Rapture: A Lightweight Gesture Recognition System from mmWave Radar Point Clouds}, 
      author={Dariush Salami and Ramin Hasibi and Sameera Palipana and Petar Popovski and Tom Michoel and Stephan Sigg},
      year={2021},
      eprint={2109.06448},
      archivePrefix={arXiv},
      primaryClass={cs.CV},
      url={https://arxiv.org/abs/2109.06448}, 
}

@article{point-xformer-nico,
   title={Point Transformer},
   volume={9},
   ISSN={2169-3536},
   url={http://dx.doi.org/10.1109/ACCESS.2021.3116304},
   DOI={10.1109/access.2021.3116304},
   journal={IEEE Access},
   publisher={Institute of Electrical and Electronics Engineers (IEEE)},
   author={Engel, Nico and Belagiannis, Vasileios and Dietmayer, Klaus},
   year={2021},
   pages={134826–134840} }

\end{document}